\pdfoutput=1

\documentclass[11pt]{article}

\usepackage[]{EMNLP2023}

\usepackage{times}
\usepackage{latexsym}

\usepackage[T1]{fontenc}

\usepackage[utf8]{inputenc}
\usepackage{multirow}
\usepackage{microtype}

\usepackage{inconsolata}
\usepackage{booktabs}
\usepackage{enumitem}
\usepackage{fdsymbol}
\usepackage{graphicx}
\usepackage{mathtools}
\usepackage{microtype}
\usepackage{multirow}
\usepackage{subfigure}
\usepackage{xspace}

\usepackage[most]{tcolorbox}
\usepackage{tikz}
\usetikzlibrary{arrows.meta} 

\newcommand{\our}{\mbox{\textsc{ToolVerifier}}\xspace}
\newcommand{\ourdata}{\mbox{\tt{ToolSelect}}\xspace}

\newcommand{\llama}{Llama-2 70B\xspace}
\newcommand{\llamachat}{Llama-2-Chat-70B\xspace}

\newcommand\blfootnote[1]{%
  \begingroup
  \renewcommand\thefootnote{}\footnote{#1}%
  \addtocounter{footnote}{-1}%
  \endgroup
}

\title{\our: Generalization to New Tools via Self-Verification}

\author{
  Dheeraj Mekala$^{*, \diamondsuit, \spadesuit}$, Jason Weston$^{\diamondsuit}$, Jack Lanchantin$^{\diamondsuit}$, \\ {\bf Roberta Raileanu}$^{\diamondsuit}$, {\bf Maria Lomeli}$^{\diamondsuit}$, {\bf Jingbo Shang}$^{\spadesuit}$, {\bf Jane Dwivedi-Yu}$^{\diamondsuit}$ \\
  $^{\diamondsuit}$ Meta~~ $^{\spadesuit}$ University of California San Diego \\
}

\begin{document}
\maketitle

\begin{abstract}


    Teaching language models to use tools is an important milestone towards building general assistants, but remains an open problem.
    While there has been significant progress on learning to use specific tools via fine-tuning, language models still struggle with learning how to robustly use new tools from only a few demonstrations.
    In this work we introduce a self-verification method which distinguishes between close candidates by self-asking contrastive questions   
    during (1) tool selection; and (2) parameter generation.
    We construct synthetic, high-quality, self-generated data for this goal using \llama, which we intend to release publicly. 
    Extensive experiments on 4 tasks from the ToolBench benchmark, consisting of 17 unseen tools, demonstrate an average improvement of 22\% over few-shot  baselines, even in scenarios where the distinctions between candidate tools are finely nuanced. 

    \blfootnote{$^*$ Work done during an internship at Meta.}
\end{abstract}
\section{Introduction}

Incorporating external tools into large language models (LLMs) enhances their real-world applicability~\cite{schick2023toolformer, Shen2023HuggingGPTSA, song2023restgpt}. 
Many tools exist in the form of APIs~\cite{Xu2023OnTT, Tang2023ToolAlpacaGT, hsieh2023tool, schick2023toolformer, Qin2023ToolLLMFL}, machine learning models~\cite{Shen2023HuggingGPTSA, Patil2023GorillaLL}, and other functions~\cite{gou2023tora}.
Nevertheless, the evolving landscape of existing tools and APIs, marked by frequent parameter updates and the daily introduction of new tools, poses a challenge for generalization. 
LLMs must quickly adapt to these changes and generalize to previously unseen tools without additional fine-tuning or extensive human input.

\begin{figure*}[t]
    \center
    \includegraphics[width=\linewidth]{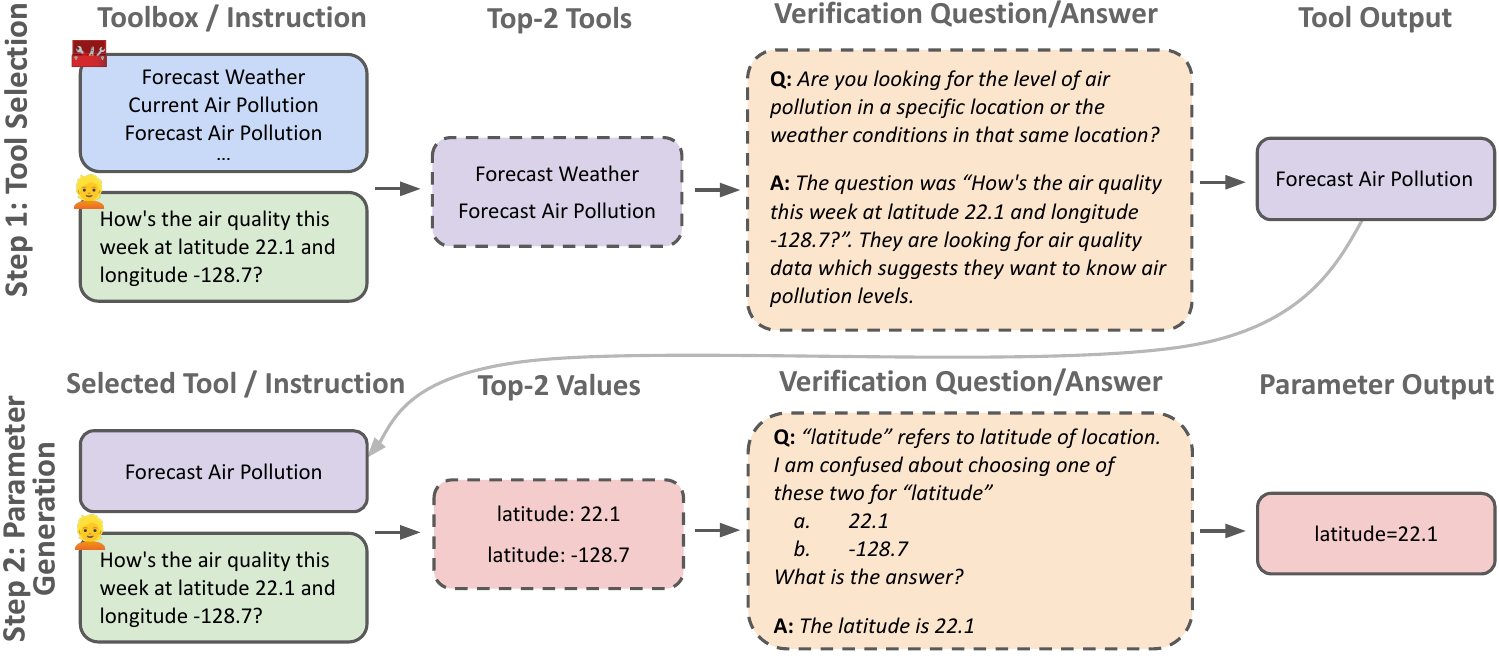}
    \caption{Overview of \our. Starting with a candidate tool list and a user instruction, \our initially identifies the top two tools. Subsequently, it generates a verification question by contrasting the selected tools and answers it. Finally, this information is appended to the context, leading to the final tool choice. The parameter generation follows a similar pipeline, wherein two candidate values are obtained for each parameter (\textit{latitude} in the above figure). Subsequently, the verification question is used to finalize the parameter value.}
    \label{fig:overview}
\end{figure*}

Several recent studies enable tool usage by fine-tuning LLMs on real~\cite{schick2023toolformer, Qin2023ToolLLMFL, Patil2023GorillaLL} or synthetic tools~\cite{Tang2023ToolAlpacaGT}, equipping them to effectively utilize tools present in the training data with a high success rate. 
Currently, the integration of unseen tools into LLMs relies on providing them with few-shot demonstrations that contain examples of user instructions and corresponding tool calls~\cite{Patil2023GorillaLL, Tang2023ToolAlpacaGT}. However, these prompting-based approaches still struggle to accurately generate a complete tool call 
from a set of unseen tools.
Moreover, this is often bottlenecked by the context length of the model, particularly when including demonstrations for a large number of tools.

To address these challenges, we propose \our, a self-verification method tailored for tool-use scenarios, capable of discerning between candidate tools and their respective parameters through verification questions.
To achieve this, we decompose the tool call generation task into two distinct sub-tasks: (1) \textit{tool selection}, 
given a user instruction, 
the most suitable tool is selected from a library 
of options,
and (2) \textit{parameter generation}, the appropriate parameters for the selected tool are then generated.
Crucially, we propose verification for each sub-task, to both improve sensitivity and to curb error propagation. Figure~\ref{fig:overview} shows an overview of each sub-task.

In the tool selection stage, our model must choose one tool among multiple options, given only the description of the tool. This stage forgoes including demonstrations, thereby significantly reducing the required context length and allowing us to select from a larger set of tools.
To facilitate learning how to choose the appropriate tool, we curate a high-quality, model-generated, synthetic training dataset containing tools, their descriptions, and user instructions.\footnote{The dataset is available at \url{https://github.com/facebookresearch/ToolVerifier}}
This dataset comprises 173 synthetic tools with corresponding descriptions, 555 samples in total, each involving reasoning about the tool's usage. We then use this dataset to fine-tune a \llama model~\cite{touvron2023llama} 
to select the correct tool for an instruction given only a set of tool names and their descriptions, allowing the model at test time to select from tools never seen during training. After the tool is selected, 
parameters are generated for the selected tool call,
which
is achieved through few-shot prompting with demonstrations corresponding to the chosen tool.

Self-verification is used at each step to reduce error propagation and enhance overall performance.
As shown in Figure~\ref{fig:overview}, for tool selection verification, we extract the top two predictions from the fine-tuned 
model. 
A verification question is then generated {\em contrasting the two options} via 0-shot prompting, enabling the model to focus on a fine-grained decision
where the answer aids in selecting one tool from the top two predictions. 
The model answers the question, and the context is updated by appending this answer to the user instruction, to guide tool selection.
A similar approach is adopted for verifying the parameter generation.

We evaluate our approach on 4 tasks from the publicly available ToolBench benchmark which tests generalization to 17 unseen real-life APIs. \our demonstrates a noteworthy 22\% improvement over few-shot prompting baselines. 
The proposed self-verification mechanism contributes an improvement of 8\%, underscoring its pivotal role  in boosting overall performance.

\section{\our}

\our chooses and calls a tool given a user instruction. It consists of the following steps:
\begin{enumerate}
\item Tool selection \& verification -- {\em selecting the tool from a library of tools}.
\item Parameter generation \& verification -- {\em generating the parameters for the tool call}.
\end{enumerate}

For step (1)  we generate synthetic data consisting of a library of tools, (instruction, tool) pairs, and reasoning notes explaining the correct choice of tool, see \autoref{tab:example_example}. Fine-tuning on this data provides improved tool selection performance, even on new sets of tools.
The selection process is then refined by verifying the choice between the top two competing choices by asking and answering {\em contrastive verification questions}, see \autoref{tab:verify_example}.

For step (2) we use few-shot prompting given demonstrations of the actual tool.
We again verify two competing likely generations.


\subsection{Tool Selection Dataset Generation}
Our first goal is to train a language model capable of selecting an appropriate tool for a given user instruction by reasoning about a candidate list of tools solely based on their names and descriptions.
We intentionally exclude demonstrations for tool selection in our approach to handle a larger set of tools in one go, using only their names and descriptions. 
In this section, we elaborate on the process of creating the training dataset for training such a tool selection language model.

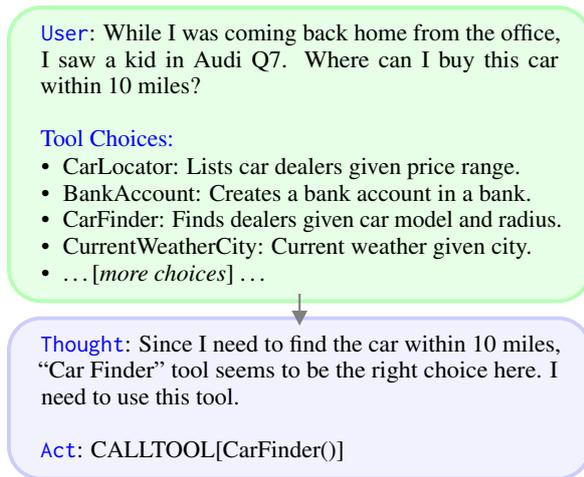
\begin{figure}[t]
\centering
\footnotesize
  \begin{tcolorbox}[colback=green!10!white, 
                    colframe=green!30!white,
                    arc=4mm,
                    auto outer arc,
                    enhanced,
                    boxsep=0pt,
                    remember as=leftbox1 
                    ]       
       \textcolor{blue}{\texttt{User}}: While I was coming back home from the office, I saw a kid in Audi Q7. Where can I buy this car within 10 miles?\\
        \\ 
        \textcolor{blue}{Tool Choices:}
       \vspace{-3mm}
        \begin{itemize}[leftmargin=*]
        \setlength{\itemsep}{0pt} 
        \setlength{\parskip}{0pt} 
            \item CarLocator: Lists car dealers given price range.
            \item BankAccount: Creates a bank account in a bank.
            \item CarFinder: Finds dealers given car model and radius.
            \item CurrentWeatherCity: Current weather given city.
            \item \dots [{\em more choices}] \dots
        \end{itemize}
  \end{tcolorbox}

  \begin{tcolorbox}[colback=blue!5!white,
                    colframe=blue!20!white,
                    arc=4mm,
                    auto outer arc,
                    enhanced,
                    boxsep=0pt,
                    remember as=leftbox2 
                    ]
               \textcolor{blue}{\texttt{Thought}}: Since I need to find the car within 10 miles, “Car Finder” tool seems to be the right choice here. I need to use this tool.\\
        \\
        \textcolor{blue}{\texttt{Act}}: CALLTOOL[CarFinder()]
  \end{tcolorbox}

\begin{tikzpicture}[overlay, remember picture]
  \draw[gray, thick, -{Triangle[length=2mm, width=2mm]}] 
    ([yshift=3mm]leftbox2.north) -- ([yshift=-3mm]leftbox1.south);
\end{tikzpicture}
\vspace{-5mm}
\caption{Illustrative training example from our synthetically constructed tool selection dataset \ourdata. Given a user instruction and a set of tools to choose from, the output consists of reasoning notes (``Thought'') and the final tool selection (``Act'').
}
\label{tab:example_example}
\end{figure}

Since the primary objective in this step is to select the correct tool (but not execute the tool call), synthetically generated tools and their corresponding descriptions can easily be used in this setting, as we do not require their actual inner workings (in order to execute them).  In our generated dataset, each training sample is thus composed of a user instruction, a candidate set of tools that includes the ground truth tool, and a reasoning note elucidating the correct choice of tool. 
An illustrative training sample is given in \autoref{tab:example_example}.

\subsubsection{Synthetic Tool Library  Generation}
\label{sec:tool_library}

\paragraph{Generation Procedure}
We generate a set of synthetic tools along with their corresponding descriptions, which are used to build the training examples.
We start by first manually annotating a ``seed set'' of eight tools and their descriptions. 
Subsequently, we employ the Llama-65B~\cite{Touvron2023LLaMAOA} model to generate additional tools using few-shot prompting with the manually annotated tools (specified in Appendix~\ref{app:tool}). 
This process then involves multiple iterations of prompting with different random seeds, where the tools generated in each iteration are integrated into the prompt for subsequent iterations to generate more diverse tools. 
Specifically, in each iteration, for every newly introduced tool, we identify the most similar tool in the prompt based on cosine similarity using RoBERTa sentence similarity~\cite{reimers2019sentence}.
We replace the most similar tool in the prompt with the new addition, ensuring a balanced diversity of tools in the prompt.
Using this iterative approach, we generate a total of 60 tools.\footnote{These tools were manually reviewed, and 7 duplicates were removed.}
It is noteworthy to highlight that this process yields a diverse set of tools from various domains including travel, banking, and calendar, with almost no manual effort.




\begin{figure}[t]
\centering
\footnotesize
  \begin{tcolorbox}[
                    colback=yellow!30!white,
                    colframe=orange!60!white,
                    arc=4mm,
                    auto outer arc,
                    enhanced,
                    boxsep=0pt,
                    remember as=leftbox1 
                    ]
I am confused to choose one of these two classes. Here are their names and descriptions:
\begin{itemize}[leftmargin=*]
\setlength{\itemsep}{0pt} 
\setlength{\parskip}{0pt} 
\item[a] CarLocator - Lists car dealers given price range.
\item[b] CarFinder: Finds dealers given car model and radius.
\end{itemize}

A contrastive question is a question that upon asking would resolve such confusion. Generate a contrastive question that I can ask myself whose answer would help me make the right choice. \\
  \end{tcolorbox}

  \begin{tcolorbox}[colback=blue!5!white,
                    colframe=blue!20!white,
                    arc=4mm,
                    auto outer arc,
                    enhanced,
                    boxsep=0pt,
                    remember as=leftbox2 
                    ]
              \textcolor{blue}{Verification Question:} What is the primary purpose of the class I need? Is it to find a car dealership based on a specific car model and location (CarFinder), or is it to list car dealerships within a given price range (CarLocator)?
  \end{tcolorbox}
\begin{tikzpicture}[overlay, remember picture]
  \draw[gray, thick, -{Triangle[length=2mm, width=2mm]}] 
    ([yshift=3mm]leftbox2.north) -- ([yshift=-3mm]leftbox1.south);
\end{tikzpicture}
\vspace{-4mm}
\caption{Verification method for tool selection: a constrastive question is generated that can then be answered to help discern among the top two predicted tools.
}
\label{tab:verify_example}
\end{figure}
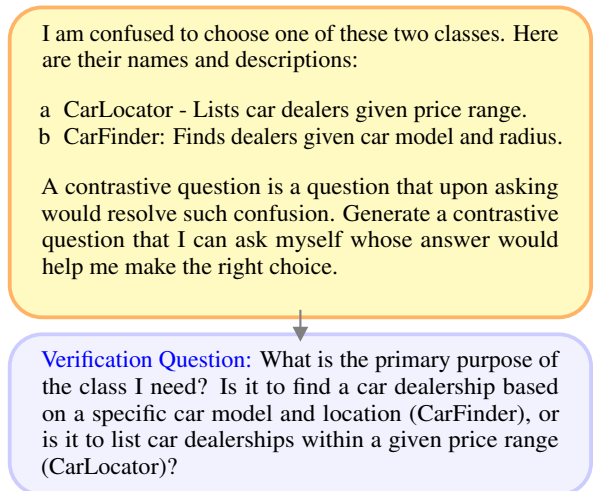

\paragraph{Generating Challenging Tool Sets}
In generating these synthetic tools, we endeavor to have a tool set that is diverse, but also sufficiently challenging. An overly simplistic training set would contain only easy choices (e.g., a weather tool versus an email tool) and this would impede the model's ability to generalize to challenging instances during test time. 
To address this, we generate two \textit{related tools} for each of our previously generated 60 tools. 
Related tools are defined as tools closely resembling a given tool but differing in either functionality or parameters. 
For instance, ``\textit{Bank account for a person name} '' and ``\textit{Bank account for an account number}'' are related tools.
We use only the tool names, and not the descriptions, for generating related tools.
After manually annotating related tools for our seed set of eight tools, we generate two related tools for each of the remaining tools with few-shot prompting with these examples, as indicated in Appendix~\ref{app:rel_tool}.

Finally, after manual inspection and curation, our dataset contains a total of 173 tools.

\subsubsection{Generating Training Examples}
Using the generated tool library, we can now generate training examples for our tool selection dataset.
This requires generating inputs (instructions), curating candidate lists of tools, and generating outputs 
(reasoning notes that explain which tools should be selected, and actions to call those tools).

\paragraph{Generating Instructions}
We first manually annotate three instructions per tool for the seed set of eight tools. Using these examples, we generate three instructions per tool for all remaining tools by few-shot prompting \llama. 

\paragraph{Curating Candidate List of Tools}
For each generated instruction, a candidate list of tools is created by 
randomly selecting 7 tools and adding the original ground truth tool for which
we generated the instruction. 
To introduce complexity, for a subset of the training set, we deliberately create challenging samples by restricting the candidate set to include only the ground truth tool and its related tools. 
This deliberate selection aims to increase the difficulty level, as distinguishing among these options is inherently more challenging than with randomly selected tools from the entire set.


\paragraph{Generating Target Outputs}
After generating the set of instructions along with their respective ground truth tool and a candidate list of tools, we create a reasoning note for each sample elucidating the rationale behind the selection of the ground truth tool, which becomes the target output for that training example (see \autoref{tab:example_example}). 
Such reasoning notes have been observed to enhance reasoning abilities~\cite{wei2022chain, yao2022react, lanchantin2023learning}.
Reasoning note generation is accomplished by prompting \llamachat with the instruction, list of tools, and the ground truth tool, and asking the model why the tool was chosen. The exact prompt used is provided in Appendix~\ref{app:prompt_config}.

Our final dataset, called \ourdata, thus contains 555 samples for our 173 tools, of which 75 samples are \textit{hard} 
 examples, featuring candidate tool sets that contain only the ground truth tool and its related tools.\footnote{The data was manually reviewed, and 56 noisy and duplicate samples were removed.}
The average number of candidate tools per instruction is 7.34 with minimum and maximum number of candidate tools being 2 and 8. 
The average length of a reasoning note is 1054 characters.
The goal of this dataset is to enable  generalization capabilities to a wide range of possible tools and tool 
libraries, and thus to demonstrate effectiveness across diverse scenarios. 

\subsubsection{Tool Selection Verification}
\label{sec:select_verify}

Despite our model being fine-tuned on the above dataset, tool selection mistakes can still happen, particularly for related tools that are hard to differentiate. Crucially, we observe that those tool selection predictions typically appear as the top few predictions -- but selection between them is challenging.  

At inference time, we thus perform the following procedure. Given an instruction:
\begin{itemize}[leftmargin=*]
        \setlength{\itemsep}{0pt} 
\item First, we use the fine-tuned tool selection model to zero-shot select a tool.
\item We then remove the initially selected tool from the candidate set of tools, and generate a second prediction. 
\item We construct a verification question to make a fine-grained decision between the model's top two selections.
\end{itemize}

We employ \llamachat to generate a contrastive verification question, where the prompt asks the model to ask a question that emphasizes the distinctions between candidate tools given their names and descriptions (see  Appendix~\ref{app:contrastive} for the exact prompt used and Figure~\ref{tab:verify_example} for an instantiation of it).
Self-asking the model regarding its predictions has been noted to reduce hallucinations~\cite{press2022measuring, Dhuliawala2023ChainofVerificationRH}, suggesting that posing such verification questions could assist the model in validating its predictions.
Since only names and descriptions are used for generating contrastive questions, they can be generated offline and utilized as needed to make the method more efficient.
The answers to these contrastive questions are obtained by further prompting \llamachat, and these are appended to the context.
Finally, we select the tool by using our fine-tuned \llama model, with the top-two tools as candidates. As the verification answer to the question is in the context this can help it select the right tool. 

\subsubsection{Parameter Generation \& Verification}

\paragraph{Parameter Generation}
Following tool selection, we generate parameters for the selected tool through few-shot prompting with \llama, utilizing demonstrations specific to the selected tool, which are assumed to be provided. Note that we do not use our synthetic tool selection dataset for parameter generation since the dataset does not contain this subtask. This procedure is only done with real tools at inference time, without prior finetuning.

\paragraph{Parameter Verification}
The generated parameters are then subjected to verification before finalizing the set, resulting in the final tool call.
To validate the generated parameters, we obtain a second set of parameter predictions.
These can be acquired using sampling or an alternative model for diverse options;  in our experiments, we employ few-shot prompting with \llamachat to obtain them.
Then, for each individual parameter, we formulate a multiple-choice question to contrast the two predictions and further prime Llama-2-Chat-70B to make a definitive choice between them, providing the parameter description and user instruction as indicated in Appendix~\ref{app:param_verification}. 
The final parameter predictions are then aggregated to construct the tool call by few-shot prompting \llama as in Appendix~\ref{app:tool_call_constr}.


\section{Experiments}

\begin{table*}[t]
    \center
    \scalebox{0.9}{
    \begin{tabular}{l c c c c | c}
        \toprule
        {\textbf{Method}} & {\textbf{Weather}} & {\textbf{Booking}}  & {\textbf{Home}} & {\textbf{Cat}} & {\textbf{Average}}\\
        \midrule
        {\em Tool-Augmented LLMs}\\
        ToolLLM 7B & $27$ & $22$ & $84$ & $26$ & $38.90$\\
        NexusRaven-V2 13B & $84$ & $93.33$ & $100$ & $98$ & $93.81$\\
        \midrule
        {\em Prompting Baselines}\\
        Single-Step \llama (1-shot)  & $79$ & $43.30$ & $100$ & $\textbf{98}$ & $78.32$\\
        Two-Step \llama (1-shot tool selection)    & $86$ & $45.00$ & $100$ & $92$ & $79.05$\\
        Two-Step \llamachat (0-shot tool selection) & $83$ & $75.80$ & $99$ & $97$ & $88.09$ \\
        \midrule
        \our (without verification)& $82$ & $98.33$ & $100$ & $96$ & $94.28$\\
        \our (tool selection verification) & $\textbf{91}$ & $\textbf{98.33}$ & $\textbf{100}$ & $97$ & $\textbf{96.67}$ \\
        \bottomrule
    \end{tabular}
    }
    \caption{{\bf Tool selection results}. We report accuracy in percentage (\%) for each task. Our fine-tuned \llama model \our, even without verification, demonstrates superior performance compared to prompting-based baselines, with a higher average performance. Our proposed tool selection verification mechanism contributes another 2.5\% improvement in accuracy on average.}
    \label{tbl:tool_selection}
\end{table*}

\begin{table*}[t]
    \center
    \scalebox{0.87}{
    \begin{tabular}{l c c c c | c}
        \toprule
        {\textbf{Method}} & {\textbf{Weather}} & {\textbf{Booking}}  & {\textbf{Home}} & {\textbf{Cat}} & {\textbf{Average}}\\
        \midrule
        {\em Tool-Augmented LLMs}\\
        ToolLLM 7B & $18$ & $0$ & $0$ & $11$ & $6.90$\\
        NexusRaven-V2 13B & $55$ & $27.50$ & $43$ & $82$ & $50.71$ \\
        \midrule
        {\em Prompting Baselines}\\
        Single-Step \llama (1-shot)   & $70$ & $7.50$ & $85$ & $83$ & $58.81$\\
        Two-Step \llama (1-shot tool selection)  & $80$ & $34.17$ & $85$ & $78$ & $67.62$\\
        Two-Step \llamachat (0-shot tool selection) & $77$ & $64.17$ & $84$ & $83$ & $76.43$ \\
        \midrule
        \our  (without verification)  & $76$ & $82.50$ & $85$ & $82$ & $81.43$\\
        \our  (tool selection verification only) & $84$ & $82.50$ & $85$ & $83$ & $83.57$ \\
        \our  (param selection verification only) & $81$ & $84.17$ & $88$ & $96$ & $87.14$ \\
        \our (tool verification+param verification) & $\textbf{90}$ & $\textbf{84.17}$ &  $\textbf{88}$ & $\textbf{97}$ & $\textbf{89.52}$ \\
        \bottomrule
    \end{tabular}
    }
    \caption{{\bf Tool call (tool selection + parameter generation) results}. We report percentage (\%) success rate for each task. Our fine-tuned \llama model \our, even without verification, results in higher performance compared to the baselines. Our proposed verification mechanism further improves the success rate by 8 points -- with both types of verification, for tool and parameter selection, each giving a separate boost in performance.}
    \label{tbl:param}
\end{table*}

In our experiments, we assess the effectiveness of our method using publicly available real-life tools.

\subsection{Tasks}
We evaluate our proposed method on four 
tool-calling tasks: Weather, Cat, Home and Booking  from ToolBench~\cite{Xu2023OnTT}.
The Weather and Cat tasks involve using the REST API to interact with the OpenWeather and Cat (images and breeds) websites, respectively. 
The Home and Booking tasks entail home search and travel booking, respectively.
The Weather, Home, and Cat tasks each comprise 100 evaluation samples, while the Booking task contains 120 samples. 
Each task includes API documentation, parameter descriptions, user instructions, and the corresponding ground truth API call pairs.

For each task, there can be more than one available tool, where the entire benchmark consists of a total of 17 tools. 
However, instead of evaluating each task individually, we make it more challenging by pooling together all available tools.
In other words, for each user instruction, the model is provided a candidate list of 17 tools.
We follow the evaluation protocol set by the benchmark and use success rate as the metric, where the success rate of a predicted tool call is 1 if its API response exactly matches the response from the ground truth API call. 

\subsection{Baselines}
We conduct a comparison with various tool-augmented LLMs and prompting baselines using \llama or \llamachat.
Specifically, for tool-augmented LLMs, we compare with ToolLLM 7B~\cite{Qin2023ToolLLMFL} and NexusRaven-V2 13B\footnote{\url{https://nexusflow.ai/blogs/ravenv2}}.
ToolLLM and NexusRaven-V2 utilize API documentation to generate tool calls corresponding to a given instruction. 
We also attempted comparing with ToolAlpaca~\cite{Tang2023ToolAlpacaGT}, however, it led to context overflow. 

For prompting baselines, we try two distinct approaches: (1) Single-step, where the model is prompted directly for an API call with a single demonstration per tool; and (2) Two-step, where we decompose the process into tool selection and parameter generation, prompting the model individually for each step, as in \our. 

The Single-step method uses 1-shot single demonstrations of each of the (17) tools to accomodate the prompt within the context size.

For the Two-step method, we consider two variants for the tool selection stage:


\begin{itemize}[leftmargin=*] 
 \setlength{\itemsep}{2pt} 
\setlength{\parskip}{0pt} 
\item {\em 0-shot:} We use a 0-shot prompt that asks to select from the list of tools, without any demonstrations for tool selection. See~\ref{0-shot-chat-llama} for the exact prompt. 
\item {\em 1-shot:} We show one demonstration per tool: a user instruction and corresponding tool name.
\end{itemize}
For parameter generation in the Two-Step method, we use three demonstrations for the selected tool. 


\subsection{\our Details and Ablations}

Our model is denoted as {\our}. For tool selection it uses  0-shot prompting with  \llama fine-tuned on our synthetic \ourdata dataset to select two tools and finalize one through our proposed contrastive-question-based tool verification. Subsequently, we generate two sets of parameters by prompting both \llama and \llamachat with three demonstrations each, and finalize the parameter set using
our proposed parameter verification. 

We additionally compare against ablated versions of our method:  with tool selection verification only (but not parameter verification), with parameter selection verification only (but not tool verification), and without verification (in either stage).






\subsection{Experimental Results}


\paragraph{Tool Selection Only}
We first report the performance of tool selection (choosing the tool correctly, but without generating parameters) in Table~\ref{tbl:tool_selection}. 
Our approach, \our, outperforms all baselines on average and individually across the majority of tasks as well.
\our performs better than all compared tool-augmented LLMs, demonstrating its superior performance.
A comparative analysis between \our with tool selection verification and without underscores the substantial enhancement in performance achieved through the verification process. 
Specifically, in tasks such as Weather and Home, we observe that the verification procedure not only improves performance in specific examples of lower baseline performance, but also does not adversely affect cases where verification may be unnecessary.

\our (both with and without verification) shows that our zero-shot \llama
fine-tuned on our synthetically generated dataset
performs better than other baselines, including a 0-shot \llamachat, with an improvement of up to 6 points. 
The average number of candidate tools per instruction in the generated training data for tool selection is 7.34 which is notably smaller than the 17 tools encountered during test time.
This difference underscores the generalization capability of our method, demonstrating its effectiveness across diverse scenarios.
It also surpasses Single-Step 1-shot and Two-Step 1-shot tool baselines by a substantial margin of more than 50 points on the challenging Booking task.
The performance of 1-shot baselines reveals the difficulty in selecting the appropriate tool from an unseen set using prompting-based approaches. In contrast, fine-tuning the model on our synthetically generated dataset with examples of using a diverse set of tools significantly improves tool selection accuracy at test time. 
Moreover, the verification procedure further improves tool selection performance by an additional 2.4 points on average.


\paragraph{Tool Call (Selection + Parameters)}
The complete tool call performance results are presented in Table ~\ref{tbl:param}. 
Similarly to the tool selection setting, \our outperforms all baselines both on average and individually across all tasks. 
\our outperforms all compared tool-augmented LLMs by a significant margin.
Comparing \our with Single-Step 1-shot highlights the challenges in generating complete tool calls at once, emphasizing the efficacy of the two-step decomposition. 
A comparative analysis between \our with and without parameter verification
illustrates that parameter verification significantly enhances performance, showing improvements of up to 14 points in the Cat task and 6 points in the Weather task, leading to an average improvement of 6 points across all tasks. 
Notably, both types of verification help, each giving a separate boost, as shown by comparing the without verification results to  tool selection verification only and tool+parameter verification.
These results underscore the significance of verification in both steps for the final tool call success.

\begin{table*}[t]
    \center
    \scalebox{0.9}{
    \begin{tabular}{l c c c c | c}
        \toprule
        {\textbf{Method}} & {\textbf{Weather}} & {\textbf{Booking}}  & {\textbf{Home}} & {\textbf{Cat}} & {\textbf{Average}}\\
        \midrule
         GPT-4$^\textbf{*}$ & $93$ & $\textbf{96.70}$ & $\textbf{97}$ & $96$ & $\textbf{95.72}$ \\
         GPT-3.5-Turbo$^\textbf{*}$ & $90$ & $85.80$ & $80$ & $92$ & $86.90$ \\
         \llama & $93$ & $84.17$ & $85$ & $86$ & $86.91$ \\
         \llamachat & $89$ & $45$ & $91$ & $88$ & $76.67$\\
        \midrule
        \our & $\textbf{99}$ & $85.80$ & $88$ & $\textbf{100}$ & $92.85$ \\
        \bottomrule
    \end{tabular}
    }
    \caption{{\bf Parameter generation results.} We report success rates (\%) in the upperbound setting where the model is provided the ground truth tool selection, and must only generate parameters. 
    We observe our fine-tuned \llama model \our outperforms  \llama and GPT-3.5-Turbo models
    in the majority of tasks and on average in this setting. Results with $*$ are taken from the Toolbench Leaderboard \citep{toolbench,Xu2023OnTT}.}
    \label{tbl:upperbound}
\end{table*}
\section{Analysis}

\subsection{Parameter Generation Only Comparison}
We additionally compare \our in the tool selection upperbound scenario, where the groundtruth tool selection is provided, and a model is only required to generate parameters through three-shot prompting. Results are given in Table~\ref{tbl:upperbound}.
\our outperforms \llamachat by 16 points as well as both \llama and GPT-3.5-Turbo by an average of 6 points on a majority of the tasks, with an improvement of up to 14 points compared to \llama in the Cat task and 8 points in the Home task compared to GPT-3.5-Turbo.
\our also demonstrates superior performance compared to GPT-4 on Weather and Cat tasks by 6 and 4 points, respectively.
This shows that our proposed method outperforms few-shot prompting approaches, even compared to stronger base models.

\subsection{Verification Question Analysis}
\paragraph{Qualitative Analysis}
Verification questions should ideally reference
the distinguishing characteristics between two given tools in order to best help the model consider the differences between the two choices. This capability is particularly crucial for closely related tools. 
For instance,  the tools "Forecast Air Pollution" and "Current Air Pollution" both provide air pollution data, but for future and current times, respectively.
Verification question generation
by \llamachat  
identifies this nuanced difference and articulates it in the verification question: \textit{Are you looking for data on the current air pollution levels in a specific location, or do you need to forecast the air pollution levels for a future date in that location?} 
Responses to such questions precisely address the identified distinction. 
An example response is: \textit{"Based on what the user said, it appears that they are looking for current air pollution data for a specific location with latitude -24.7 and longitude -57.3. Therefore, the answer is: A. Retrieve current air pollution data for a specific location."} 
Inserting this response into the context for model prediction, amplifies the selection of the appropriate tool, guiding the model towards the correct choice.

For more distinct tools,  the model captures  higher level differences. For example, for "Forecast Air Pollution" and "Get favorite cat images", the generated question inquires about the user's interests: \textit{Which aspect are you more interested in: predicting environmental air quality or exploring feline visuals?}

\paragraph{Significance of Contrastive Questions}
To demonstrate the significance of contrastive-question-based verification, we conduct an experiment by zero-shot prompting \llamachat to choose one tool from the top-2 without employing a verification question. 
Instead, we present the names and descriptions of the top-2 tools and frame it as a multiple-choice question, asking \llamachat to make a selection. 
The prompt is mentioned in Appendix~\ref{sig:contra}. 
We experiment on the Weather task and the accuracy of \llamachat is 70\% whereas the accuracy of contrastive question-based verification is 91 \%.
This significant enhancement over straightforward prompting illustrates the effectiveness of contrastive questions.

\paragraph{Instruction-Conditioned Verification}

In our proposed approach we generate verification questions using solely the names and descriptions of the top-2 selected tools, see \autoref{tab:verify_example}. 
We can compare this to conditioning on the user instruction as well, by adding it to the prompt.
Conditioning on the instruction during verification still shows improvement over the no-verification baseline (89 versus 82). However, prompting with the user instruction slightly decreases performance compared to the non-user-conditioned verification, dropping accuracy from 91 to 89, perhaps because the  decision is biased to be more similar to the original top choice being verified, which was also based on the instruction. Note if the tool set is not too large, using only names and descriptions has the benefit that the questions can be precomputed and cached. 

\begin{figure}[t]
    \centering
    \includegraphics[width=0.48\textwidth]
    {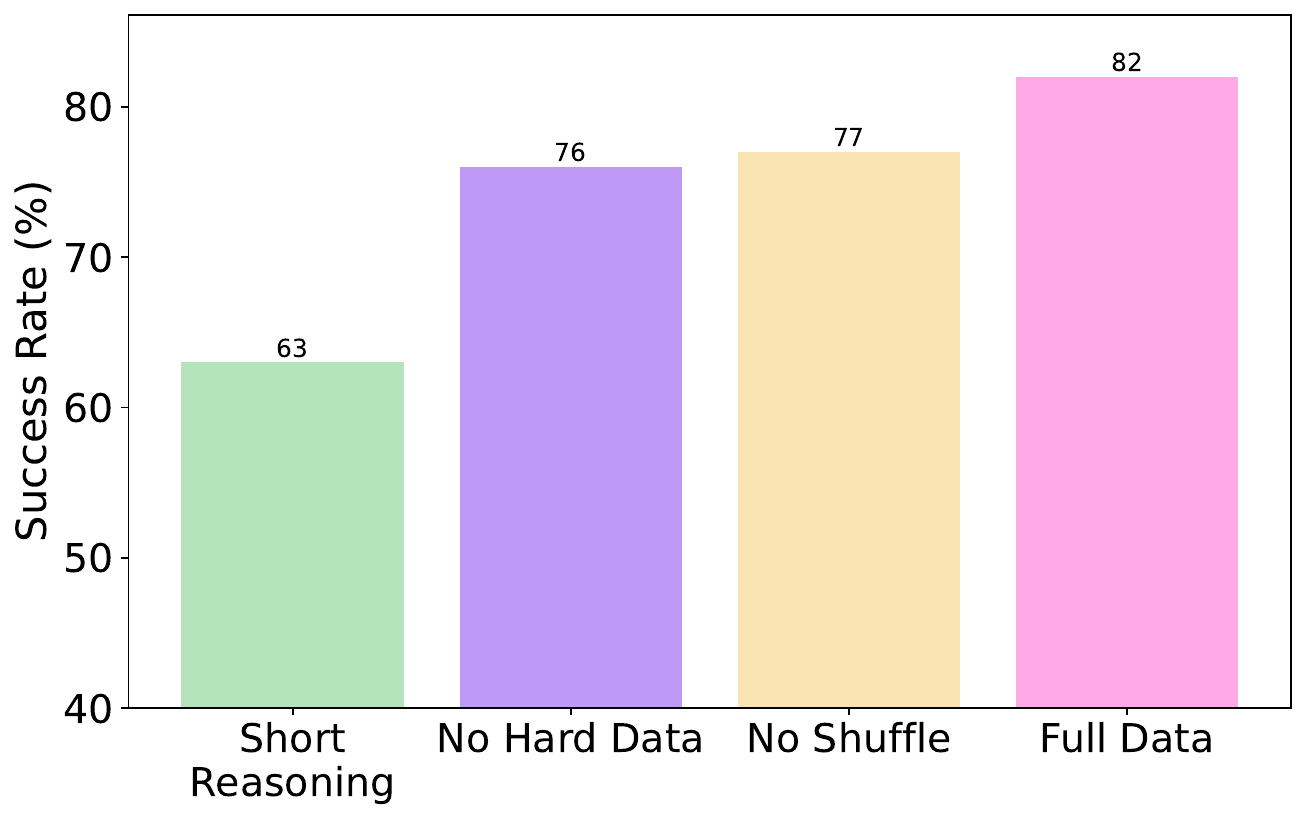}
    \caption{We analyze various aspects of our synthetic 
    \ourdata training data including the ordering of the candidate tool list (``No Shuffle''), 
    difficulty level (''No Hard Data''), and the length of reasoning notes (``Short Reasoning''). We find samples with longer reasoning notes, difficult samples, and randomly ordered candidate tool lists contribute to high performance (``Full Data'').
    }
    \label{fig:perf_training_data}
\end{figure}

\subsection{Synthetic Training Data Analysis}






We analyze  our
synthetic \ourdata training data through various ablations, with results on tool selection for the Weather task given in Figure~\ref{fig:perf_training_data}.

Challenging training samples (samples that have a candidate tool list containing related tools to the ground truth tool, see \autoref{sec:tool_library}) are found to improve generalization.
To assess the impact of these challenging samples, we remove them and train a model solely with easier samples (``No Hard Data''). 
The results indicate a notable 6-point drop in performance after excluding the hard samples, highlighting their significance.

Next, we experiment by reducing the maximum reasoning note length from 480 tokens to 200 tokens (``Short Reasoning'') and observe a significant drop in performance, up to 19 points. 
Shorter reasoning texts are significantly less helpful in guiding appropriate tool selection.

Lastly, we compare performance with different orderings of the candidate tool list. 
In the ``No Shuffle'' scenario, the ground truth tool in the training data is always positioned first. Implementing this ordering strategy results in a 5-point drop in performance, underscoring the significance of randomly shuffling the candidate tool list in the training data.

\subsection{Parameter Verification Error Analysis}
In the parameter verification step, we identify a consistent pattern in errors while answering the verification questions, 
predominantly involving common sense errors where the model tends to hallucinate values instead of adhering to the user instruction, which is also observed in \citet{mekala-etal-2023-zerotop}. 
A notable example of such errors occurs with the \textit{min-price} parameter in Booking tool, which signifies the minimum price the user is willing to pay for a booking.
In 5 instances out of 19 wrong predictions for the Booking task, when the user specifies only their maximum budget, the model generates the maximum value for the min-price parameter rather than 0. 
Similar errors are observed with the \textit{min-area} parameter in the Home task. 
In 4 instances out of 12 mistakes, when the user expresses the desire for a home given only a maximum area, the model incorrectly predicts the mentioned value as the minimum, instead of using 0.

\section{Related Work}

\paragraph{Self-Verification}

Outside the domain of tool use, iterative improvement of  the generation capabilities of LLMs
typically involves prompting an LLM to provide feedback on given generated facts or answers and subsequently refining their outputs~\cite{Madaan2023SelfRefineIR, Shridhar2023SCREWSAM} which has also been shown to reduce hallucination~\cite{Dhuliawala2023ChainofVerificationRH}. 
Additionally, some studies involve the fine-tuning of custom LLMs to better accommodate feedback~\cite{Yu2023TeachingLM, Shridhar2023TheAO, Zhang2023SelfConvincedPF}, aiming to enhance reasoning in chain-of-thought prompting for improved downstream performance. 
~\citet{Lu2023SELFLS} also focus on fine-tuning LLMs to effectively incorporate and act upon such iterative feedback. 
In this paper, we focus on tool usage, whereas previous works typically focus on generation. 
We propose a novel contrastive verification method, specifically for tool selection and parameter generation.
Our approach contrasts the choice between selecting options, whereas previous work typically verifies single facts or answers for generation of responses. 

\paragraph{Enabling Tool Use in LLMs}

Many approaches have emerged for enabling tool usage in LLMs,
involving techniques such as few-shot prompting with tool-use demonstrations across diverse tool categories, including APIs~\cite{Qin2023ToolLLMFL, Chen2023ChatCoTTC}, machine learning models~\cite{Shen2023HuggingGPTSA, Patil2023GorillaLL}, code interpreters~\cite{Gao2022PALPL, Chen2022ProgramOT}, and mathematical functions~\cite{gou2023tora}. Additionally, several approaches advocate for fine-tuning LLMs on custom-generated datasets tailored for tool usage~\cite{schick2023toolformer, Tang2023ToolAlpacaGT, Parisi2022TALMTA, Xu2023OnTT, Patil2023GorillaLL, srinivasan2023nexusraven}.  \citet{Yang2023GPT4ToolsTL} propose fine-tuning on multi-modal tools. Furthermore, recent works introduce auxiliary sources such as tool documentation~\cite{hsieh2023tool} and tool tokens~\cite{Hao2023ToolkenGPTAF} to facilitate tool usage.
Despite the plethora of works focused on enabling tool usage in LLMs, to the best of our knowledge none has explored verification methods for this purpose. This paper aims to fill this gap by introducing multi-step contrastive verification.

\paragraph{LLMs for Data Generation}
LMs have been used for generating training data for various tasks including text classification~\cite{mekala-etal-2021-coarse2fine, mekala-etal-2022-leveraging}, semantic similarity~\cite{Schick2021GeneratingDW}, and instruction tuning~\cite{Wang2022SelfInstructAL, Honovich2022UnnaturalIT, Xu2023BaizeAO, alpaca}. 
Several works ~\cite{Tang2023ToolAlpacaGT, Qin2023ToolLLMFL, Tang2023ToolAlpacaGT, schick2023toolformer, Patil2023GorillaLL,srinivasan2023nexusraven} have employed LLMs to generate synthetic tools or tool use examples.
In this paper, we generate synthetic training data 
to support our self-verification procedure.
\section{Conclusion}


In this paper, we present a generation and self-verification method for 
enhancing the performance of tool calls for LLMs. 
This involves decomposing the tool call generation task into two distinct sub-tasks: tool selection and parameter generation, where we apply verification at each step.
Model-generated verification questions allow nuanced decision-making between related tools, helping it to correct mistakes.
Experimental results on four tasks from the publicly available ToolBench benchmark demonstrate substantial improvements using our approach.

\section{Limitations}
Our self-generated verification questions and answers are produced in a zero-shot manner, making them effective for general-purpose tools but may necessitate further training for niche tools. Additionally, our framework is currently designed for single-tool-usage tasks and does not support instructions requiring multiple or compositional tool usage.

\section{Ethics Statement}

This paper introduces a self-verification method for tool calling that generates verification questions to aid in making accurate choices with confidence. As such, we do not expect that the fine-tuning self-verification process should introduce biases not already observed in the model, and we do not anticipate any significant additional ethical concerns beyond those issues already seen in standard systems \citep{weidinger2021ethical}.

\section{Acknowledgments}
The authors thank Omer Levy and Chunting Zhou for helpful discussions.

\bibliography{custom, anthology}

\newpage
\appendix

\clearpage\newpage
\newpage
\section{Appendix}

\subsection{Prompts \& Configurations}
We use top-p sampling while generating with a temperature set to $0.7$.

\subsubsection{Tool Generation}
\label{app:tool}
The prompt for tool generation using few-shot prompting LLaMa-65B is:
\vspace{-0.1cm}
\begin{center}
\begin{tcolorbox}[colback=green!10!white, 
                    colframe=green!30!white,
                    arc=4mm,
                    auto outer arc,
                    enhanced,
                    boxsep=0pt,
                    remember as=leftbox1 
                    ]       
    \small
Name: Humidity\\
Description: Computes humidity at a location on a date\\

Name: Trip Booking\\
Description: Makes a travel booking\\

Name: Currency Conversion\\
Description: Converts an amount from one currency to another.\\

Name: Age Calculator\\
Description: Calculates the age based on a given birthdate and the current date.\\

Name: Search Engine\\
Description: Searches online about a query\\

Name: Restaurant Finder\\
Description: The Restaurant Finder tool finds the restaurants based on its location, cuisine and the number of people.\\

Name: Movie Review\\
Description: The Movie Review tool gets top-rated movie reviews for a particular movie.\\

Name: Pizza Order\\
Description: The Pizza Order tool orders a pizza with provided toppings and size.\\

Name:
  \end{tcolorbox}
\end{center}

\subsubsection{Reasoning Note Generation}
\label{app:prompt_config}
The prompt for reasoning note generation using \llamachat is:
\vspace{-0.1cm}
\begin{center}
\begin{tcolorbox}[colback=green!10!white, 
                    colframe=green!30!white,
                    arc=4mm,
                    auto outer arc,
                    enhanced,
                    boxsep=0pt,
                    remember as=leftbox1 
                    ]       
        \small
        [INST] <<SYS>>\\
You are a helpful assistant.\\
<</SYS>>\\

Here are the list of available tools:\\
\{Candidate tool list\} \\

A user said, "\{instruction\}". \\

To answer this, you found Tool "\{name\}" to be the most suitable than other tools. Why?[/INST] \\
  \end{tcolorbox}
\end{center}

In the above prompt ``\{instruction\}'' denotes the user instruction and ``\{name\}'' denotes the ground truth tool. ``\{Candidate tool list\}'' contains names and descriptions of each tool.

\subsubsection{Related Tools Generation}
\label{app:rel_tool}
The prompt for related tool generation using few-shot prompted \llama is:
\vspace{-0.1cm}
\begin{center}
\begin{tcolorbox}[colback=green!10!white, 
                    colframe=green!30!white,
                    arc=4mm,
                    auto outer arc,
                    enhanced,
                    boxsep=0pt,
                    remember as=leftbox1 
                    ]       
    \small
Name1: Humidity\\
Name2: Humidity at timezone\\
Name3: Humidity Altitude Location date\\

Name1: Book Review\\
Name2: Book Review By Date\\
Name3: Book Review By Day\\

Name1: Car Rental\\
Name2:  Car Rental with insurance\\
Name3: Car Rental with driver\\

Name1: \{name\}\\
Name2:
  \end{tcolorbox}
\end{center}
In the above prompt \{name\} denotes the name of the tool whose related tools are being generated.
While generating multiple related tools per original tool, we generate one related tool after another with different seeds, to improve the diversity of the related tools.

\subsubsection{Contrastive Question Generation}
\label{app:contrastive}
\begin{center}
\begin{tcolorbox}[colback=green!10!white, 
                    colframe=green!30!white,
                    arc=4mm,
                    auto outer arc,
                    enhanced,
                    boxsep=0pt,
                    remember as=leftbox1 
                    ]       
   \small
        [INST] <<SYS>>\\
You are a helpful assistant.\\
<</SYS>>\\

I am confused to choose one of these two classes. Here are their names and descriptions:\\
a. \{name1\} - \{description1\}\\
b. \{name2\} - \{description2\}\\

A contrastive question is a question that upon asking would resolve such confusion. Generate a contrastive question that I can ask myself whose answer would help me make the right choice.[/INST] \\
  \end{tcolorbox}
\end{center}
In the above prompt ``\{name1\}'', ``\{description1\}'' and ``\{name2\}'', ``\{description2\}'' are names and descriptions of two selected tools respectively.

\subsubsection{Parameter Verification}
\label{app:param_verification}
\begin{center}
\begin{tcolorbox}[colback=green!10!white, 
                    colframe=green!30!white,
                    arc=4mm,
                    auto outer arc,
                    enhanced,
                    boxsep=0pt,
                    remember as=leftbox1 
                    ]       
        \small
        [INST] <<SYS>>\\
You are a helpful assistant.\\
<</SYS>>\\

A user said, "\{instruction\}"\\

{parameter definition}\\

For the above user instruction, I am confused about choosing one of these two for "\{parameter name\}".\\
a. \{prediction 1\}\\
b. \{prediction 2\}\\

What is the answer? Answer the following question strictly based on what the user said above. If there is no mention, respond with "None". If there is, select the answer from the given options and respond with the chosen option only in square brackets []. [/INST] \\
  \end{tcolorbox}
\end{center}
In the above prompt ``\{instruction\}'' denotes the user instruction.
``\{parameter name\}'' represents the parameter name under verification. 
Additionally, ``\{prediction 1\}'', ``\{prediction 2\}'' signify two parameter predictions obtained from \llama and \llamachat, respectively.

\subsubsection{0-shot Chat LLaMa-70B}
\label{0-shot-chat-llama}
\begin{center}
\begin{tcolorbox}[colback=green!10!white, 
                    colframe=green!30!white,
                    arc=4mm,
                    auto outer arc,
                    enhanced,
                    boxsep=0pt,
                    remember as=leftbox1 
                    ]       
        \small
        [INST] <<SYS>>\\
You are a helpful assistant.\\
<</SYS>>\\

Here are the list of available tools:\\

\{Candidate tool list\}\\

A user said, "\{instruction\}"\\

What tool to use for the above instruction? Respond with just the name of the tool[/INST] \\
  \end{tcolorbox}
\end{center}

In the above prompt ``\{instruction\}'' denotes the user instruction and ``\{Candidate tool list\}'' contains names and descriptions of each tool.

\subsubsection{Tool Call Construction}
\label{app:tool_call_constr}
\begin{center}
\begin{tcolorbox}[colback=green!10!white, 
                    colframe=green!30!white,
                    arc=4mm,
                    auto outer arc,
                    enhanced,
                    boxsep=0pt,
                    remember as=leftbox1 
                    ]       
        \small
        INS: A user says, "Please retrieve the temperature, humidity, wind, and visibility data at place with latitude = -37.3, longitute = 1.9."\\
lat: -37.3\\
lon: 1.9\\
units: none\\
mode: none\\
lang: none\\
API: curl -X GET 'https://api.openweathermap.org/data/2.5/weather?lat=-37.3\&lon=1.9\&appid={API\textunderscore KEY}\&units=none\&\\mode=none\&lang=none'\\

INS: A user says, "How is the weather now in location with longitute 125.9 and latitude 39.0? Respond in simplified Chinese with json format and imperial units."\\
lat: 39.0\\
lon: 125.9\\
units: imperial\\
mode: json\\
lang: zh\textunderscore cn\\
API: curl -X GET 'https://api.openweathermap.org/data/2.5/weather?\\lat=39.0\&lon=125.9\&appid={API\textunderscore KEY}\&units=imperial\\\&mode=json\&lang=zh\textunderscore cn' \\

INS: A user says, "Give me a current weather report for place where longitute is 174.4 and latitude is -19.0."\\
lat: -19.0\\
lon: 174.4\\
units: none\\
mode: none\\
lang: none\\
API: curl -X GET 'https://api.openweathermap.org/data/2.5/weather?lat=-19.0\&lon=174.4\&appid={API\textunderscore KEY}\&units=none\\\&mode=none\&lang=none'\\

INS: A user says, "\{instruction\}"\\
\{param\textunderscore str\}\\
API:
  \end{tcolorbox}
\end{center}

In the above prompt ``\{instruction\}'' denotes the user instruction and ``\{param\textunderscore str\}'' contains parameters and their predicted values.

\subsubsection{Significance of Contrastive Questions}
\label{sig:contra}
An example prompt is provided below. 
\begin{center}
\begin{tcolorbox}[colback=green!10!white, 
                    colframe=green!30!white,
                    arc=4mm,
                    auto outer arc,
                    enhanced,
                    boxsep=0pt,
                    remember as=leftbox1 
                    ]       
        \small
        [INST] <<SYS>>\\
You are a helpful assistant.\\
<</SYS>>\\

A user says, "Please retrieve the temperature, humidity, wind, and visibility data for next week with latitude = -37.3, longitute = 1.9."\\

To address the above instruction which one of the below tools is the most suitable? Select the answer from the given options and respond with the chosen option ONLY in square brackets []. \\

A. Forecast Air Pollution = Get the future air pollution data in location with latitude=\{lat\}, longitude=\{lon\}\\
B. Forecast Weather Latitude Longitude = Get the weather data for future in location with latitude=\{lat\}, longitude=\{lon\}[/INST]\\
  \end{tcolorbox}
\end{center}

\subsection{Hyperparameters for \llama Fine-tuning}
\label{app:hyp}
We fine-tune \llama for 3 epochs with a learning rate of 1e-5 with warm up. The effective batch size is 8 and the weight decay is 0.1.
We train it on 16 A100 GPUs.

\subsection{Frequently Asked Questions}

\paragraph{Why did you use LLaMa-65B for tool generation instead of \llama?}
The 70B model was not released by the time we generated tools. Hence, we used the available 65B model.

\section{API Details}

The four APIs pertaining to ToolBench are the Weather, Booking, Home, and Cat APIs. To execute the API calls, we registered for access to the Weather and Cat API, whereas for Home and Booking we ensured correct syntax, as proposed in the benchmark~\cite{toolbench}.

\end{document}